\pdfoutput=1

\documentclass[11pt,final]{article}

\usepackage{acl}
\usepackage{times}
\usepackage{latexsym}
\usepackage[T1]{fontenc}
\usepackage[utf8]{inputenc}
\usepackage{microtype}
\usepackage{inconsolata}
\usepackage{graphicx}

\title{Are Multilingual Language Models an Off-ramp for Under-resourced Languages? Will we arrive at Digital Language Equality in Europe in 2030?}

\author{Georg Rehm \\
  DFKI GmbH, Germany\\
  Humboldt-Universität zu Berlin, Germany\\
  \texttt{georg.rehm@dfki.de} \small (corresponding) \\
    \\\And
    Annika Grützner-Zahn \\
  DFKI GmbH, Germany\\
   \\\And
    Fabio Barth \\
  DFKI GmbH, Germany\\
  }

\begin{document}

\maketitle

\begin{abstract}
Large language models (LLMs) demonstrate unprecedented capabilities and define the state of the art for almost all natural language processing (NLP) tasks and also for essentially all Language Technology (LT) applications. LLMs can only be trained for languages for which a sufficient amount of pre-training data is available, effectively excluding many languages that are typically characterised as \emph{under-resourced}. However, there is both circumstantial and empirical evidence that multilingual LLMs, which have been trained using data sets that cover multiple languages (including under-resourced ones), do exhibit strong capabilities for some of these under-resourced languages. Eventually, this approach may have the potential to be a technological off-ramp for those under-resourced languages for which ``native'' LLMs -- and LLM-based technologies -- cannot be developed due to a lack of training data. This paper, which concentrates on European languages, examines this idea, analyses the current situation in terms of technology support and summarises related work. The article concludes by focusing on the key open questions that need to be answered for the approach to be put into practice in a systematic way.
\end{abstract}

\section{Introduction}

Especially in today's data-driven, machine learning and language model-based era of Language Technologies (LTs), it is intuitively evident that some languages have better, more advanced, more sophisticated technology support than others. This intuitive notion concerns not only the scope and deployment of technologies in applications or devices but also the coverage and robustness of these technologies. The natural language processing literature typically distinguishes between \emph{high-resource} and \emph{low-resource}, i.\,e., \emph{under-resourced}, languages. Concentrating on the languages of Europe and the European Union (EU), we are faced with a multilingual society with 24 official EU Member State languages and more than 60 additional languages (regional and minority languages, co-official languages etc.) plus languages of immigrants, trade partners and also tourists. For decades, most  European languages have been considered under-resourced, which is peculiar since all EU Member State languages are politically on equal footing according to the EU Treaty. However, technically they are clearly not -- this fact has been officially recognised by the European Union \citep{EPresolution2018} in a resolution that calls for the development of technologies and resources for \emph{all} European languages, supported through a large-scale and long-term funding programme.

Concurrently with the publication of the EP resolution \citep{EPresolution2018} and the subsequent EU project European Language Equality (ELE, see Section~\ref{sec:ele}), the full potential and groundbreaking capabilities of large language models (LLMs) have become clear to the natural language processing (NLP) and Artificial Intelligence (AI) research community and also to the public at large with a number of products such as, most notably, ChatGPT, released in November 2022. Through the development of multilingual LLMs, which have been trained on data sets covering not only one but multiple languages, there is growing evidence that such multilingual LLMs exhibit strong capabilities for some or many of the under-resourced languages they have been trained on even though only very little training data for these languages was actually included in the pre-training data set. If this does in fact hold for many or all under-resourced languages, this technical approach could become a technical off-ramp for these under-resourced languages for which our field is unable to develop ``native'' stand-alone technologies simply because there is too little pre-training data available. This off-ramp would not only avoid digital language extinction and digital language death, it would also bring us closer to digital language equality in Europe, maybe even in time by 2030, as initially demanded by the ELE project.

This position paper attempts to examine the question if multilingual LLMs can be considered an off-ramp for under-resourced languages with a special emphasis on European languages. First, Section~\ref{sec:ele} introduces the concept of digital language equality including the current state of technology support of Europe's languages. Section~\ref{sec:data} concentrates on the available data including a look at the current situation and recent developments how to improve it. Section~\ref{sec:offramp} introduces the idea of making use of multilingual LLMs, trained on both high as well as low-resource languages, as an off-ramp for low-resource languages. Section~\ref{sec:summary} concludes the article.

\section{Digital Language Inequality in Europe}
\label{sec:ele}

In Europe’s multilingual setup, all 24 official EU languages are granted equal status by the EU Charter and the Treaty on EU. The EU is also home to over 60 regional and minority languages which have been protected and promoted under the European Charter for Regional or Minority Languages (ECRML) treaty since 1992, in addition to various sign languages and the languages of immigrants as well as trade partners. Artificial Intelligence, Natural Language Processing, Natural Language Understanding, Language Technologies, and Speech Technologies have the potential to enable multilingualism -- and the multilingual European information society -- technologically but, as the META-NET White Paper Series \emph{Europe’s Languages in the Digital Age} \citep{rehm2012} found already in 2012, our languages suffer from an extreme imbalance in terms of technological support: English is very well supported through technologies, tools, data sets and corpora, but languages such as, among others, Maltese, Estonian and Icelandic have hardly any support at all. In fact, the 2012 study assessed ``at least 21 European languages to be in danger of digital extinction''. If, as mentioned above, all European languages are supposed to be on an equal footing in general, technologically, they clearly are not \citep{kornai2013}.

After the META-NET findings and several follow-up projects, studies and recommendations \citep[e.\,g.,]{rehm2013h,STOA2017}, the joint CULT/ITRE report \emph{Language Equality in the Digital Age} \citep{EPresolution2018} was eventually passed by the European Parliament with an overwhelming majority in September 2018. It concerns the improvement of the institutional framework for LT policies at the EU level, EU research and education policies to improve the future of LTs in Europe, and the extension of the benefits of LTs for both private companies and public bodies. This EP resolution also recognises that there is an imbalance in terms of technology support of Europe’s languages, that there has been a substantial amount of progress in research and technology development and that a large-scale, long-term funding programme should be established to ensure full technology support for all of Europe’s languages. The goal is to enable multilingualism technologically since “the EU and its institutions have a duty to enhance, promote and uphold linguistic diversity in Europe” \citep{EPresolution2018}.

While the resolution was an important milestone, there has been no concrete follow-up action along the lines laid out in the resolution, i.\,e., to set up “a large-scale, long-term coordinated funding programme for research, development and innovation in the field of language technologies, at European, national and regional levels, tailored specifically to Europe’s needs and demands” \citep{EPresolution2018}. In the meantime, however, many highly influential breakthroughs in the area of language-centric AI have been achieved, mostly by large enterprises in the US and Asia, especially approaches and technologies concerning large language models (LLMs). 

The EU project European Language Equality (ELE), which ran from 2021 until 2023, set out to analyse the current technology support of Europe's languages ten years after the META-NET study \citep[see][for a comprehensive overview of this project's results]{rehm2023}. It defined digital language equality (DLE) as the ``state of affairs in which all languages have the technological support and situational context necessary for them to continue to exist and to prosper as living languages in the digital age.'' \citep[][p.~43]{gaspari2023}. The results of the ELE project demonstrate that we are still very far away from this ideal state: \emph{except for English and, to a certain extent, French, Spanish and German, all European languages must be considered significantly or even massively under-resourced when comparing their individual scores on the Digital Language Equality Metric} \citep{gaspari2022,gruetzner-zahn2022}. In other words, the findings from the META-NET study and the EP resolution have still been valid in 2023 and they continue to be valid today. As a direct result of these observations, the ELE project prepared strategic recommendations \citep{rehm2023b}, presented to the European Union, the full implementation of which would improve the situation so that many of the currently under-resourced languages would eventually benefit from the development of novel and more language technologies as well as from making available novel and more data sets.\footnote{Despite promising initial discussions with the European Commission and the European Parliament in 2023, at the current point it does not seem particularly likely that the ELE Programme will be financed by the European Union.} 

\begin{figure*}[tb]
\centering
\includegraphics[width=\textwidth]{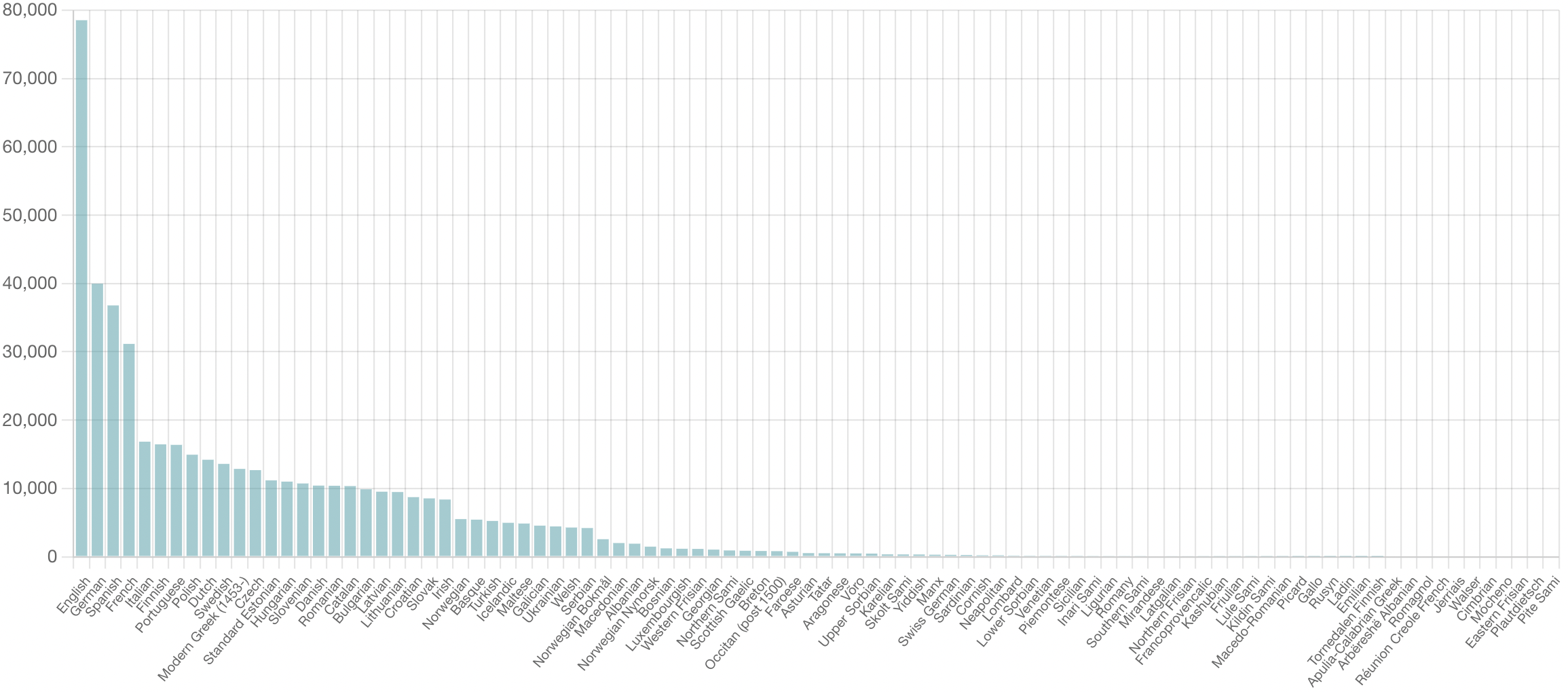}
\caption{The technological scores of the Digital Language Equality Metric (as of February 2025): the numbers shown in the bar chart strongly correlate with the data available for each language}
\label{fig:dashboard}
\end{figure*}

The DLE Metric is ``a measure that reflects the digital readiness of a language and its contribution to the state of technology-enabled multilingualism, tracking its progress towards the goal of DLE.'' \citep[][p.~43]{gaspari2023}, see \citet{gaspari2022} and \citet{gruetzner-zahn2022} for more details. One implementation of the DLE Metric is available in the European Language Grid (ELG) platform \citep{rehm2022a} in the form of a  dashboard.\footnote{\url{https://live.european-language-grid.eu/catalogue/dashboard}} Figure~\ref{fig:dashboard} shows the current state (as of mid February 2025) of technology support of Europe's languages according to the DLE dashboard; the empirical basis is the set of approx.~18,000 language resources and language technologies available in ELG. The figure shows that English is the best supported language by far, followed by German, Spanish and French. The long tail essentially starts with Italian, Finnish and Portuguese, followed by Polish, Dutch and Swedish.

While there are several very promising developments in Europe that will all contribute to improved digital language equality (see the following sections), it is also very much evident that a lack of available data for Europe's languages is one of the main current bottlenecks that we are facing.

\section{Availability of European Language Data: Current State and Future Developments}
\label{sec:data}

Regarding this situation, the most important, relevant and promising development both in the European Union and also across various European countries relates to \emph{data spaces}. The EU has been implementing its Data Strategy since 2019.\footnote{\url{https://commission.europa.eu/strategy-and-policy/priorities-2019-2024/europe-fit-digital-age/european-data-strategy_en}} A key component of the industrially-oriented single market for data, in which ``data can flow within the EU and across sectors, for the benefit of all'', ``European rules [\dots] are fully respected'' and in which ``the rules for access and use of data are fair, practical and clear'', is the vision of establishing ``interoperable data spaces'' for ``pooling European data in key sectors''. The \citet{ecstaffdoc2022} describes how these data spaces are supposed to be set up and operated including relevant legislation \citep[also see][]{opendei_2021}. This document also lists a number of `official' EU data spaces targeting sectors such as manufacturing, mobility, health, finance, energy, agriculture and skills. The Common European Language Data Space \citep[LDS,][]{rehm2024a}, funded through the Digital Europe Programme (DEP),\footnote{\url{https://digital-strategy.ec.europa.eu/en/activities/digital-programme}} is one of these official EU data spaces.\footnote{\url{https://language-data-space.ec.europa.eu}} 

The LDS project develops a decentralised digital infrastructure for the sharing of any kind of language data (written text, spoken language, multimodal data etc.). Due to the importance of LLMs for industrial and academic applications and their dependence on vast amounts of language data for training models, especially when considering the increasing push of the EU towards digital sovereignty, the development of the LDS has recently become increasingly relevant. The LDS is one crucial part of a bigger initiative towards more European independence and increased European participation in the global LLM landscape (including research, development, innovation, application, deployment and monetisation). 

The main principle in the concept of data spaces is that of data sovereignty, i.\,e., the ability of a person, natural or legal, to exclusively decide, in a sovereign way, on the usage of their own data as an economic asset. As a result of this principle, the LDS gives owners/providers full control regarding access to their data, also enabling data transactions including monetary transactions. The LDS will support tracking of data provenance and lineage and it will enforce access and usage policies, formally codified as contracts established in a trustful environment. The typical operation technically supported in LDS is that of exchange of metadata about data assets (i.\,e., data, data products and data services) and exchange or transfer of such assets between trusted participants. An asset can be any type of language-related data set (corpus, lexicon etc.) or a language processing service. The LDS will eventually be a pan-European marketplace allowing users, providers and consumers to perform, manage and monitor their commercial transactions and exchange data in full respect of the contractual agreements made between the parties.

All participants, data providers and data consumers, have to install the LDS connector software through which they can participate in this marketplace. It enables the documentation of language data, cataloging of assets by the connector's owner, contract-based agreement for the exchange of data, transfer or processing of data and logging of all transactions. The overall marketplace can be conceptualised as the aggregation of all its connectors, connected in a peer-to-peer fashion. Each connector is equipped with a local catalogue in which this provider’s offerings are published and available for querying or crawling by other connectors. In addition, a central catalogue will give an overview of all offerings publicly available. 

The LDS project has recently concluded its second year and it has also recently made available Version 1.0.0 beta of the LDS connector software, which is now being tested. The LDS user group, which was started in 2024, already has representatives of more than 100 organisations, all of which are interested in sharing or using language data. With the LDS infrastructure, Europe will have a robust and sustainable mechanism for sharing any kind of language data in a  systematic and also trustworthy way. The LDS will have a very strong positive impact on the overall language data situation in Europe including the availability of data for under-resourced languages and making available language data from various industries and sectors.

However, it will take a few years for this initiative to become so active, vibrant and successful that thousands of participants from all over Europe, representing all European languages (including minority, co-official, regional and other languages), are comfortably represented in this marketplace, which is why we need to examine short-term alternatives for developing technologies that all European languages can benefit from so that they do not experience digital language death \citep{kornai2013}.

\section{Multilingual LLMs as an Off-ramp for Under-resourced Languages}
\label{sec:offramp}




Multilingual LLMs have demonstrated remarkable cross-linguistic generalisation and zero-shot transfer learning capabilities, significantly advancing NLP and LT applications. It must be stressed, though, that most multilingual LLMs are trained on relatively small data sets for most languages, except for English. Still, a number of these multilingual LLMs perform well on benchmarks assessing the LLM's capabilities on these under-resourced languages and they also outperform monolingual LLMs trained on these under-resourced models only due to cross-lingual transfer \citep{chang-etal-2024-multilinguality}. Leaderboards, like the European LLM Leaderboard,\footnote{\url{https://huggingface.co/spaces/openGPT-X/european-llm-leaderboard}} show for instance that LLama \citep{llama2, llama3}, while only having a small amount of training data on low-resource languages like Romanian, Portuguese, Slovenian or Slovakian, reaches high accuracy scores for evaluation tasks on language understanding. It primarily focuses on eight languages but has demonstrated strong zero-shot and few-shot capabilities across a broader range \cite{multiEUbenchmarks}. Another relevant case is the model Gemma 2 \citep{team_gemma_2024} which is stated to be trained mainly on English data, yet it also achieves surprisingly high accuracy scores for a number of European low-resource languages, like Slovenian, Slovakian, Latvian and Romanian.

Several European multilingual LLMs have been developed especially with the idea to support European languages, including under-resourced ones, in a more systematic way. The models that exhibit strong transfer capabilities to low-resource languages are BLOOM \citep{BLOOM}, Mistral Nemo\footnote{\url{https://huggingface.co/mistralai/Mistral-Nemo-Instruct-2407}, \url{https://mistral.ai/en/news/mistral-nemo}}, EuroLLM \citep{EuroLLM}, Teuken-7B \cite{teuken} and Salamandra \cite{salamandra2025}. BLOOM was trained on data covering 46 natural languages, Mistral Nemo on nine languages, Salamandra and EuroLLM on the official EU-24 plus eleven additional languages, and Teuken-7B \cite{teuken} on the EU-24 languages. All of these models demonstrate consistent performance across all included languages \cite{multiEUbenchmarks}. The models also show that LLMs trained on more equally distributed multilingual training data sets are able to compete with the models trained on predominantly English data and in some cases even outperform them \cite{multiEUbenchmarks}, i.\,e., multilingual pre-training can yield strong results even for languages with limited direct training data.

These promising and astonishing results have led to our idea of using pre-trained multilingual LLMs as off-ramps for under-resourced languages. They show that it may be possible to develop LLMs with enough linguistic capabilities on low-resource languages (see Figure~\ref{fig:dashboard}), so that these languages can be actually used in the digital sphere by their native speakers and so that the corresponding LLMs can be used to develop and provide technologies for these language communities. For this to become a realistic vision, several factors and current limitations need to be examined more closely.

First, the availability and quality of data impacts significantly which languages can be covered by a pre-trained model. Currently, there are a handful of sufficiently large multilingual data sets for pre-training. Data sets like the ROOTS corpus \citep{roots}, MADLAD-400 \citep{madlad}, CulturaX \citep{culturax}, mC4 \citep{mt5}, HPLT \citep{HPLT} and various Wikipedia dumps are commonly used. MADLAD-400 covers the most languages (419) while pointing to aspects regarding data quality for under-resourced languages. In addition, except for ROOTS, all data sets are based on CommonCrawl dumps. Even ROOTS, although trying to gather data from other sources, had to complement their data with a subset of OSCAR, which is also based on CommonCrawl. These limitations in usable data sources and the assumption of ``the more, the better'' lead to using as much data as possible for under-resourced languages without considering the question how much under-resourced language data is actually needed.

There have been initial approaches to limit the training of LLMs by restricting the size of vocabulary and data following the idea of language learning in human development with promising results \citep{warstadt_findings_2023, georges_gabriel_charpentier_not_2023}. Moreover, more systematic testing found the relevance of the size of the training data set for under-resourced languages for multilingual LLMs up to a certain point depending on the capacity of the model. Other factors influencing cross-linguistic transfer are linguistic similarity and script type \citep{chang-etal-2024-multilinguality, ahuja-etal-2023-mega, bagheri-nezhad-agrawal-2024-drives}. As these findings are only indicative and taking into account the complexity of the topic, a careful, systematic testing and analysis of the influence of LLM training data composition with a special emphasis on analysing the resulting capabilities regarding under-resourced languages is still missing. This includes aspects like the number of languages per model or proportionally per size, the impact of typological similarity or diversity of the languages, which language serves as the high-resource language from which the model transfers the linguistic capabilities, to name just a few.

In addition to the training data, the training process itself can be adapted to increase the efficiency of the learning process. Bootstrapping word translation pairs from monolingual corpora \citep{improving_low} can improve intrinsic cross-lingual word representation quality and downstream task performance in under-resourced languages of multilingual LLMs. In-context learning methods \citep{llms_are_low_resource}, layer-wise data sampling \citep{pan_lisa_2024} or increased architecture flexibility \citep{georges_gabriel_charpentier_not_2023} are additional ways for improving the performance without the need for more training data. Combining these training-efficient methods with the questions about necessary training data quantity and data set composition for under-resourced languages can yield effective answers for using multilingual LLMs as off-ramps for under-resourced languages.

Finally, benchmarking is crucial in evaluating multilingual language models, but existing evaluation frameworks are often very much English-centric and culturally biased toward western or US culture \cite{Tao_2024}. The widely used GlobalMMLU benchmark \cite{globalMMLU} was created by automatically translating the established MMLU \cite{mmlu} benchmark into 42 languages and manually annotating the data set to verify and culturally de-bias the benchmark. It was annotated through a community effort to identify culturally-sensitive samples. However, \citet{globalMMLU} showed that the languages were not annotated in a balanced way. Also, determining if a data set is truly culturally inclusive is critical, as simple translations can create a false sense of inclusivity without addressing deeper cultural biases. Especially in terms of following up on the off-ramp idea, the goal should be to incorporate diverse cultural knowledge to ensure fairness in multilingual NLP/AI evaluation \cite{globalMMLU}. A counter-example is the INCLUDE \cite{include} benchmark, developed to assess multilingual language understanding with a focus on regional knowledge, ensuring a more nuanced evaluation of multilingual LLM performance and incorporating cultural knowledge.

\section{Summary and Conclusions}
\label{sec:summary}

The \emph{Strategic Agenda for Digital Language Equality} \cite{rehm2023b}, developed by the European project ELE, outlines a number of steps, instruments and mechanisms in terms of how to achieve the overall goal of arriving at digital language equality in Europe. As the full implementation of the complex and expensive ELE programme currently seems unlikely, we developed the rather pragmatic and short-term term idea to ``exploit'' multilingual LLMs as off-ramps for under-resourced European languages. This way, speakers of these languages would be able to benefit from sophisticated, state of the art language technologies, enabling them to participate in the digital sphere in their mother tongues. 

While there are still several open gaps in research (see Section~\ref{sec:offramp} and also below), the overall situation in Europe can be perceived as promising. The EU and also the Member States provide bigger budgets for AI and LLM research and deployment projects and also for overarching, pan-European initiatives such as the AI Factories and the Alliance for Language Technologies EDIC (ALT-EDIC), they also concentrate on the development of data spaces including, crucially, the Common European Language Data Space. In the next five to seven years we can expect more and more participants in the LDS who will actively share, sell and purchase larger amounts of written or spoken language data, especially for developing LLMs. In this regard, Europe needs to make an effort to make available especially those language data sets that do exist but that are currently simply not available for LLM development purposes. Obivous candidates are news data, as produced on a daily basis by newspapers and online portals, and also audio-visual data, as produced, also on a daily basis, by public and private broadcasters throughout Europe, covering essentially all relevant European languages. It will be a crucial challenge to ``unlock'' these classes of potential LLM training data for developing multilingual LLMs that are tailor-made for Europe's languages \citep[see][for a suggestion how to put this idea into practice]{rehm2023c}.

Before the technological shortcut can be systematically used, a number of open research questions need to be addressed: How much under-resourced language data do we need for the multilingual LLM to exhibit strong capabilities for this language? Are there specific constraints regarding the high-resource languages that also need to be include in the multilingual LLM? Should these be, for example, typologically similar to the under-resourced language(s) or is using English sufficient? Are there alternatives to using English as the predominant base language of a multilingual pre-training data set? How many languages should be included in the pre-training data in total? Does a diverse set of languages improve multilingual performance? 

\section*{Acknowledgments}

The work presented in this paper has received funding from the German Federal Ministry for Economic Affairs and Climate Action (BMWK) through the project OpenGPT-X (project no.~68GX21007D). The Common European Language Data Space is funded by the European Union through the contract LC-01936389.

\bibliography{main}

\end{document}